# Reliable and superior elliptic Fourier descriptor normalization and its application software ElliShape with efficient image processing

Hui Wu[1, 2, 3, 4#], Jia-Jie Yang[1, 3, 4#], Chao-Qun Li[5], Jin-Hua Ran[2, 4, 6], Ren-Hua Peng[6, 7]∗ and Xiao-Quan Wang[1, 2, 3, 4, 6]∗

[1] Big Data and AI Biodiversity Conservation Research Center, Institute of Botany, Chinese Academy of Sciences, Beijing 100093, China

[2] State Key Laboratory of Plant Diversity and Specialty Crops and Key Laboratory of Systematic and Evolutionary Botany, Institute of Botany, Chinese Academy of Sciences, Beijing 100093, China

[3] Plant Science Data Center, Chinese Academy of Sciences, Beijing, 100093, China

[4] China National Botanical Garden, Beijing 100093, China

[5] School of Life Sciences, Qilu Normal University, Jinan, 250200, China

[6] University of Chinese Academy of Sciences, Beijing 100049, China

[7] Key Laboratory of Noise and Vibration Control, Institute of Acoustics, Chinese Academy of Sciences, Beijing 100190, China

# Equally contributed to this work.

**Corresponding authors:**

    **Prof. Dr. Xiao-Quan Wang**
    Institute of Botany, Chinese Academy of Sciences
    Haidan, Beijing 100093, China
    E-mail: xiaoq_wang@ibcas.ac.cn

    **Associate Prof. Dr. Ren-Hua Peng**
    Institute of Acoustics, Chinese Academy of Sciences,
    Haidian, Beijing 100190, China
    E-mail: pengrenhua@mail.ioa.ac.cn

# Abstract

1. Elliptic Fourier analysis (EFA) is a powerful tool for shape analysis, which is often employed in geometric morphometrics. However, the normalization of elliptic Fourier descriptors has persistently posed challenges in obtaining unique results in basic contour transformations, requiring extensive manual alignment. Additionally, contemporary contour/outline extraction methods often struggle to handle complex digital images.

2. Here, we reformulated the procedure of EFDs calculation to improve computational efficiency and introduced a novel approach for EFD normalization, termed "true EFD normalization", which remains invariant under all basic contour transformations. These improvements are crucial for processing large sets of contour curves collected from different platforms with varying transformations. Based on these improvements, we developed ElliShape, a user-friendly software. Particularly, the improved contour/outline extraction employs an interactive approach that combines automatic contour generation for efficiency with manual correction for essential modifications and refinements.

3. We evaluated ElliShape's stability, robustness, and ease of use by comparing it with existing software using standard datasets. ElliShape consistently produced reliable reconstructed shapes and normalized EFD values across different contours and transformations, and it demonstrated superior performance in visualization and efficient processing of various digital images for contour analysis.

4. The output annotated images and EFDs could be utilized in deep learning-based data training, thereby advancing artificial intelligence in botany and offering innovative solutions for critical challenges in biodiversity conservation, species classification, ecosystem function assessment, and related critical issues.

**Code and data:** The source codes and data are deposited at https://www.plantplus.cn/ElliShape.

# 1. Introduction

Morphological variations, both within and between species, occur pervasively throughout the life on earth. Accurately capturing and comprehensively analyzing biological morphology and its variations are crucial for understanding their interconnections with developmental and environmental forces. Geometric morphometric (GM) offers mathematical tools for quantitative analysis of multi-dimensional biological forms, which was thought to be a revolution in morphometrics (Rohlf & Marcus, 1993; Adams et al., 2004). Several GM methods have been developed using coordinates of landmarks or curves, such as landmark and outline methods (Adams et al., 2004; Xu & Bassel, 2020; Mitteroecker et al., 2022). Landmark-based GM methods excels in analyzing morphologies using a set of biologically definable points that is assumed to be homologous across samples, but are less effective for shapes with few definable points, like leaves and petals (Adams et al., 2004). In contrast, outline-based GM methods capture the bounding edge of biological shapes and analyze shapes using mathematical functions.

Elliptic Fourier analysis (EFA), initially proposed by Kuhl and Giardina in 1982 for object recognition in images, extracts arbitrary two-dimensional close curves into elliptic Fourier descriptors (EFDs) with a mathematical function, and synthesizes the original shape with any minor allowable deviations . This method is a powerful tool for quantitative shape analysis, classification, reconstruction, and modeling, which has been widely used in plant science, marine biology, evolutionary biology, and anthropology (Bucksch et al., 2017; Caple et al. 2017; MacLeod & Kolska Horwitz, 2020). With the recent deep learning advancements for image segmentation, it's worthily to expect that integrating EFA with these large segmentation model could significantly advance the study of biological morphology, in which most of the biology morphology information are collected in specimen images. Over the past two centuries, natural history collections have amassed millions of specimens, many of which are now digitalized and accessible via platforms like GBIF (www.gbif.org), iDigBio (www.idigbio.org), and CVH (www.cvh.ac.cn). This vast dataset offers a valuable opportunity to leverage artificial intelligence for accelerating innovations in biodiversity studies (Davis, 2023). However, a significant hurdle in handling these images is the scarcity of organ-specific datasets with detailed annotation (e.g., shape, texture) across diverse taxa, which could potentially hinder the application of deep learning techniques (Hussein et al., 2022). Addressing this challenge is essential for advancing our understanding of biological diversity and morphology.

The EFA-based GM analysis procedure involves contour/outline extraction, EFDs calculation, and EFD normalization. Contour/outline extraction obtains the chain code, a concise encoding of a shape's contour using a series of directional changes, from an image segment for EFDs calculation. The EFD normalization eliminates scale, rotation, translation, symmetry, and other basic transformations to ensure a unique representation of the original form. However, Kuhl and Giardina (1982) did not provide detailed steps for obtaining the chain code, and their procedures for EFD normalization are limited to scale, rotation and translation, which have persisted as a challenge in the field (Bonhomme et al., 2014; Wishkerman & Hamilton, 2018). Over the past several decades, numerous methods and software tools have been developed for implementing EFA, including DiaOutline, MASS, Momocs, SHAPE, ShapeR, and SHERPA (Iwata & Ukai, 2002; Kloster et al., 2014; Bonhomme et al., 2014; Libungan & Pálsson, 2015; Wishkerman & Hamilton, 2018; Chuanromanee et al., 2019). These software tools facilitate contour/outline extraction via user-friendly graphical user interfaces (GUI) and often require manually alignment to improve EFD normalization when it fails to give uniform results under basic transformations. For example, the R package Momocs provides alignment option using control points and normalizes EFDs in terms of size and rotation using the "first ellipse" corresponding to the selected control points. However, this alignment process is quite unproducible, especially when dealing with a large number of shapes.

Additionally, practical challenges arise in contour/outline extraction. DiaOutline, SHERPA, and ShapeR are specifically designed for extracting contours/outlines in diatom taxa (single cells, a rectangular-like shape) and fish. MASS, a MATLAB software, requires images with a white background for EFA. Despite recent advancements in image processing, resulting in automated extraction tools like GinJinn2 (Ott & Lautenschlager, 2022) and Segment Anything (Kirillov et al., 2023), automatic outline extraction from complicated specimen images, especially with large, realistic contour datasets, remains challenging (Hussein et al., 2022). Therefore, it is of great significance to improve the approach for true EFD normalization and develop a user-friendly software for organ annotation and geometric feature extraction from digital images.

In this study, we improved the entire EFA-based GM morphological analysis process. First, we introduced an efficient EFDs calculation method. Second, the true EFD normalization across all basic contour transformations are presented. Overall, we introduced ElliShape to simplify contour/outline extraction, improve EFDs calculation efficiency, and achieve true EFDs normalization.

## 2. Improved approach for elliptic Fourier descriptors normalization

This section introduces an improved approach for obtaining normalized Fourier descriptors from a chain-encoded contour. The key steps for calculating true normalized elliptic Fourier descriptors are described below.

### 2.1 Efficient elliptic Fourier descriptors calculation

Inspired by Kuhl and Giardina (1982), we reformulated elliptic Fourier descriptors (EFDs) for concise and efficient computation. Contours (outlines) are represented using chain code, composed of basis vector line segments $\{0,1,2,\mathrm{K},7\}$ as shown in Appendix S1. The chain code starts from the origin point on the $xy$ plane through all the line segments step-by-step.

Suppose $\{A_0, C_0\}$ and $\{a_n, b_n, c_n, d_n\}$ are the direct components (DC), and the $n$-th order EFDs for a given chain code $V$. Detail steps for calculating the EFDs from the chain code $V$ can be found in the original paper by Kuhl and Giardina (1982). We reformulate the DC components as follows:

$$A_0 = \frac{1}{2T}\sum_{p=1}^{K}(x_p + x_{p-1})\Delta t_p \tag{1}$$

$$C_0 = \frac{1}{2T}\sum_{p=1}^{K}(y_p + y_{p-1})\Delta t_p \tag{2}$$

where $T$ is the basic period of the chain code, and $K$ is the total number of basis line segments. $x_p$ and $y_p$ are the projections on the $x$ and $y$ axis of the first $p$ links of the chain code, respectively. $\Delta t_p$ is the time needed to traverse the $p$-th link. It should be noted that the DC components $\{A_0, C_0\}$ presented in this paper are equivalent to those presented in the original paper but have a much simpler expression, improving computational efficiency.

The original $n$-th order EFDs: $a_n$, $b_n$, $c_n$ and $d_n$ are calculated as follows:

$$a_n = \frac{T}{2n^2\pi^2}\sum_{p=1}^{K}\left(\frac{\Delta x_p}{\Delta t_p}\left(\cos\left(\frac{2n\pi t_p}{T}\right) - \cos\left(\frac{2n\pi t_{p-1}}{T}\right)\right)\right) \tag{3}$$

$$b_n = \frac{T}{2n^2\pi^2}\sum_{p=1}^{K}\left(\frac{\Delta x_p}{\Delta t_p}\left(\sin\left(\frac{2n\pi t_p}{T}\right) - \sin\left(\frac{2n\pi t_{p-1}}{T}\right)\right)\right) \tag{4}$$

$$c_n = \frac{T}{2n^2\pi^2}\sum_{p=1}^{K}\left(\frac{\Delta y_p}{\Delta t_p}\left(\cos\left(\frac{2n\pi t_p}{T}\right) - \cos\left(\frac{2n\pi t_{p-1}}{T}\right)\right)\right) \tag{5}$$

$$d_n = \frac{T}{2n^2\pi^2}\sum_{p=1}^{K}\left(\frac{\Delta y_p}{\Delta t_p}\left(\sin\left(\frac{2n\pi t_p}{T}\right) - \sin\left(\frac{2n\pi t_{p-1}}{T}\right)\right)\right) \tag{6}$$

We can reconstruct the original curve using the first $N$ orders EFDs which were shown in Appendix S1 under $N=1$, $N=5$, $N=15$ and $N=35$, respectively.

**2.2 True elliptic Fourier descriptors normalization**

The aim of true elliptic Fourier descriptors normalization is to find a unique set of descriptors corresponding to a specific chain code or, more generally, a specific close-form contour curve $(x(t), y(t))$, regardless of planar translation, scaling, rotation, reversal, different starting points, or symmetric placement of the contour. To address these transformations, we propose a comprehensive procedure for obtaining true normalized elliptic Fourier descriptors, which is independent of the mentioned basic curve transformations (Fig. 1). This is crucial when dealing with numerous contour curves collected from various platforms.

To obtain a unique set of descriptors corresponding to a specific chain code, we present the overall normalization procedure for obtaining true normalized EFDs step-by-step as follows:

(a) Set the DC to zero for **planar translation normalization**;
(b) Adjust the heading direction of the 1st order elliptic curve for **anti-clockwise**;
We can use the cross production $\Omega$ to detect the heading direction as follows:

$$\begin{cases} \Omega < 0 \rightarrow clockwise \\ \Omega > 0 \rightarrow anticlockwise \end{cases} \quad (7)$$

where $\Omega = (a_1, c_1) \otimes (b_1, d_1) = a_1 d_1 - c_1 b_1$ and $\otimes$ is the cross-production operator.

(c) Use the major axis length of the 1st order EFD for **scale normalization**, which is calculated as follows:

$$D(\theta_t^*) = \sqrt{(a_1^2 + c_1^2)\cos^2(\theta_t^*) + (a_1 b_1 + c_1 d_1)\sin(2\theta_t^*) + (b_1^2 + d_1^2)\sin^2(\theta_t^*)} \quad (8)$$

where $\theta_t^*$ is the included angle between the major axis and the $x$ axis, which is obtained as follows:

$$\begin{cases} \theta_t^* = \theta_t, if\ D(\theta_t) \geq D(\theta_t + \pi/2) \\ \theta_t^* = \theta_t + \dfrac{\pi}{2}, else \end{cases} \quad (9)$$

where $\theta_t$ is defined as:

$$\theta_t = \frac{1}{2}\arctan\left(\left|\frac{2(a_1 b_1 + c_1 d_1)}{a_1^2 + c_1^2 - b_1^2 - d_1^2}\right|\right) \quad (10)$$

(d) Rotate the major axis of the 1st order elliptic curve to the x-axis in clockwise direction for **rotation normalization** after scale normalization.

(e) Choose the endpoint of the major axis of the 1st order elliptic curve as the starting point to do the **starting point normalization** after rotation normalization, and fix the starting point of the 1st order elliptic curve to the positive unit vector on the x-axis after starting point normalization.

(f) Adjust the sign value for x-symmetric and y-**symmetric normalization** according to the 2nd order elliptic Fourier coefficients.

### 3. Python software ElliShape

To facilitate the improved EFA-based morphological analysis, we first provided a series of Python scripts to ensure better cross-platform compatibility. Subsequently, we introduced ElliShape, a user-friendly software that facilitates simple outline extraction, efficient EFD calculation, and true EFD normalization. The user manual for ElliShape is presented in Appendix S2.

**Overall design.** The software consists of two parts: contour/outline extraction and EFA, as illustrated in Fig. 2. The contour/outline extraction module combines automatic contour generation for efficiency, and allows for polygon selection and manual corrections to ensure accurate modifications and refinements. This process involves four steps: target selection, segmentation, contour outlining after manual correction, and chain code generation (Fig. 2A-C).

The segmentation step utilizes the Otsu thresholding method (Suzuki, 1985) or the Segment Anything Model (SAM) algorithm (Kirillov et al, 2023) to automatically generate a binary image with a bright foreground (object) region and a dark background region, facilitating subsequent contour outlining. The contour outlining step integrates Suzuki, Canny, Sobel, Prewitt, Roberts, log, and zero crossing algorithms as alternative options to obtain the chain code. After chain code generation, clicking the "elliptic Fourier analysis" button transitions to a new section (Fig. 2D) for acquiring true normalized EFDs and visualizing reconstructed shapes.

**Key function:**

**Object selection.** The software incorporates the SAM model to automatically derive object boundaries by clicking on the center point of the target object, which can improve the target selection to a new level. We also retained the "Polygon Tool" function, allowing for manual creation of polygonal regions, which is crucial for selecting overlapping objects (Fig. 2A) and in cases where the SAM model fails to perform its task.

**Segmentation and chain code generation.** We have integrated functions for comprehensive image contour/outline extraction generated in the preservation and digitization of specimens, addressing issues such as incomplete edges due to white strips, low image contrast, and various types of noise from fragile, overlapping, and damaged organs. These specific functions are listed as follows:

"*Inverted colors*" and "*Image enhancement*". The automatic segmentation process assumes higher grayscale values and noticeable contrast of foreground objects against the background. For black background images, the "*Inverted colors*" function should be applied before object selection to meet these conditions, as shown in Fig. 3A. Following object selection, "*Image enhancement*" can be used to improve boundary contrast in low contrast images, as demonstrated in Fig. 3B.

"*Erosion*" and "*Dilation*". The erosion operation (removing pixels from the boundaries of objects), as illustrated in Fig. 3C, eliminates background noises that interferes with the desired object. Meanwhile, the dilation operation (adding pixels to their contour of objects), demonstrated in Fig. 3D, smoothens small gaps or holes in the target region to obtain contiguous curves.

"*Editing window*". The manual editing window (Fig. 2B) enables users to mend edges by dragging the left mouse button to deal with the white strips, as demonstrated in Fig. 3B. White Strips, commonly used to mount specimens, can cause fragmented contours and persist as an obstacle for outline extraction compared to other software.

"*Chain Code*". This function converts edge coordinates into chain code as a preprocessing step for EFDs calculation. The binarization window (Fig. 2B) shows the object contour (green line) and the reconstructed shape using the obtained chain code (red line). In cases of chain code interruptions, corresponding positions will be concurrently showcased in the Editing window for easy modifications.

**Size calculation**. The software supports area and circumference calculation by selecting two points on a ruler, if available, and inputting the corresponding realistic distance to determine the actual dimensions of individual pixel points. The results, including boundary coordinates, encoded chain code, area and perimeter sizes, and labeled images, are saved with user-defined tags (Fig. 2C). Emphasizing the importance of labeled images from diverse taxa, these contribute significantly to building large datasets, thereby accelerating deep learning-based biodiversity research for scientific discovery.

**Elliptic Fourier analysis.** This section comprises two functions: EFD normalization and visualization of reconstructed shapes (Fig. 2D). Users can define the harmonic number of EFD and select normalization options. True EFD normalization is achieved when all normalization options are chosen. In the visual window, the left pane shows shapes reconstructed using EFDs (red line) and chain-code (blue line), while the right pane displays reconstructed shapes using normalized EFDs.

## 4. Comparisons with existing methods

We conducted a comprehensive comparative analysis to assess the stability and robustness of EFD normalization and to evaluate the ease of use in contour/outline extraction. We compared the original procedures proposed by Kuhl and Giardina (1982) and four representative software applications: SHAPE (Iwata & Ukai, 2002) , MASS (Chuanromanee et al., 2019), Momocs (Bonhomme et al., 2014), and DiaOutline (Wishkerman & Hamilton, 2018). SHAPE and MASS are comprehensive computer program packages including functions for contour/outline extraction and EFA analysis. Momocs provides basic EFDs calculation and EFD normalization based on boundary coordinates. DiaOutline is specifically designed for contour/outline extraction.

**4.1 Comparisons of EFD normalization**

Six representative graphics were chosen as standard examples (Fig. 4A), and their chain codes were subjected to basic graphic transformations to assess the robustness of the candidate normalization methods. All the transformed chain codes were normalized using the original procedures, SHAPE, Momocs, MASS and our newly proposed procedure, respectively. Detailed comparisons are presented in Appendix S3. Notably, only our proposed procedure succeeded in obtaining a unique set of normalized EFDs under all basic transformations, demonstrating its superiority over existing methods and software.

In Fig. 4B, the turtle form, indicated as P1, along with its eight basic transformation forms, was presented. The reconstructed shapes, utilizing normalized harmonic coefficients of EFDs from both the original and our procedures, are shown in Fig. 4C. Our method consistently produced identical results across all conditions, as evidenced by the normalized harmonic coefficients (Fig. 4E). This observation was further confirmed with additional examples detailed in Appendix S4.

The original procedure was constraint to normalize scales, rotation, and translation under specific circumstances (Fig. 4C, D). In SHAPE, the EFDs were normalized based on the first harmonic ellipse or the longest radius/ the farthest point on the contour from its centroid to ensure invariance to scales, rotation, translation, and starting point (although not universally effective). EFD calculation in MASS and Mocmocs is based on the original procedure and SHAPE, respectively, primarily designed to improve usability through a graphical user interface. Therefore, none of these methods/software could yield a unique set of normalized EFDs corresponding to a specific chain code with all basic types of transformations simultaneously.

Additionally, we utilized 319 leaves from three Fagaceae species (*Castanea dentata*-97 leaves, *Fagus sylvatica*-93 leaves and *Quercus alba*-129 leaves) extracted from digitized herbarium specimens to further evaluate the stability and robustness of our approach. These leaves exhibited three outline shapes: elongated ellipses without lobes, ellipses without lobes, and forms with lobes. Each leaf's outline chain code underwent transformation using the eight basic transformations mentioned above, resulting in a total of 2871 chain codes. We employed principal component analysis (PCA) on the normalized EFDs to reduce dimensionality while retaining principal components, providing a visual representation for comparison. Extracting the first two dimensions from the 140-dimensional normalized EFDs, shown in Fig. 5, revealed a concentrated distribution for the three shapes, despite PC1 and PC2 explaining less than 50% of the variance. This suggests a notable improvement in the normalization of leaf shapes with our new procedure. Similar analyses were conducted on *Pinus* cones, as shown in Appendix S5, further validating our findings.

**4.2 Comparison of contour/outline extraction**

In our software package, a collection of representative images (located in the "Sample" folder of the source codes) was employed to comprehensively evaluate performance in target selection and edge segmentation. Our software demonstrated superior performance in visualization and rapid processing of various digital images (Fig. 3). In contrast, existing software struggled to handle all samples (Appendix S6).

SHAPE is limited by its use of rectangle frames for target acquisition and lacks editing function, restricting its application, particularly for damaged organs during outline extraction. Additionally, results verification in SHAPE must be conducted in subsequent steps. Notably, SHAPE has not been updated since 2006 and fails when dealing with lengthy chain codes (Appendix S6 A-H). DiaOutline simplifies automatic target selection based on user-defined size criteria, often encompassing numerous non-target shapes (Appendix S6 I-P), and lacks further refinement and editing capabilities. MASS allows manual creation of polygonal regions but is restricted to extracting shapes against a white background (Appendix S6 Q-S).

**5. Conclusions and perspectives**

This paper introduces a rigorous mathematical procedure for achieving true EFD normalization, ensuring invariance with planar translation, scaling, rotation, reversal, and symmetry, which is crucial for the analysis of massive and complex contours. Additionally, we present ElliShape, a user-friendly software package designed to enhance contour/outline extraction. ElliShape provides an integrated platform with functionalities including image enhancement, contour/outline extraction, EFA analysis with the proposed true EFD normalization, and labeled data collection.

We have made the Python scripts for ElliShape publicly available, and we have developed an online version (https://www.plantplus.cn/ElliShape) to promote global engagement in specimen exploration. Our ongoing improvements aim to enhance its data processing capabilities. Our ultimate goal is to establish a large-scale specimens dataset using ElliShape, which has the potential to significantly advance artificial intelligence (AI) applications in botany. This initiative promises to deliver innovative insights and solutions in biodiversity conservation, species classification, and ecosystem function assessment.

**Conflict of interest**

The authors declare no conflict of interest.

**Author contributions**

H.W., R.H.P. and X.Q.W. designed the study, R.H.P. and H.W. improved the approach for elliptic Fourier descriptors normalization, J.J.Y., H.W. and R.H.P. developed software ElliShape, J.J.Y. and H.W. collected the data, H.W., R.H.P. and J.J.Y. wrote the manuscript, X.Q.W., J.H.R and C.Q.L. revised the manuscript. All authors gave final approval for publication.

**Data accessibility**

The source codes and data are deposited at https://www.plantplus.cn/ElliShape.

**Acknowledgments**

This study was supported by National Key Research and Development Program of China (Grant No. 2022YFF130170), the Informatization Plan of Chinese Academy of Sciences (Grant No. CAS-WX2021SF), and the Youth Innovation Promotion Association, CAS (Grant No. 2022023).

**Figures**

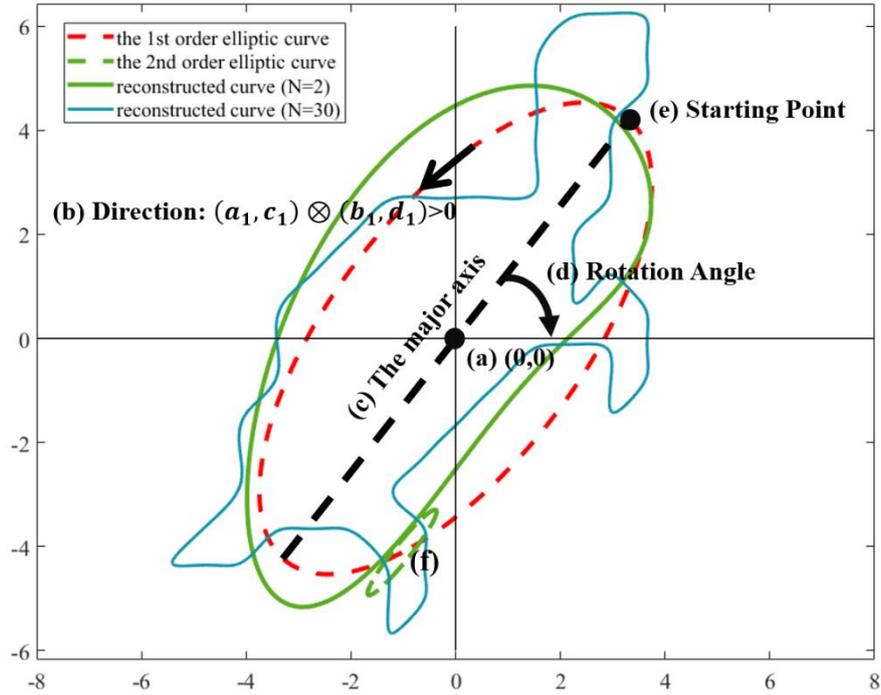

Figure 1. The procedure of true elliptic Fourier descriptor normalization, using the graphics of a turtle (blue line) as an example: Firstly, set the direct components to zero for planar translation normalization (a). Secondly, we determine the heading direction using right hand rule and normalize the heading direction to anti-clock wise (b) based on the 1st order elliptic curve (red dotted line) of the original curve. Thirdly, we use the major axis length of the 1st order elliptic curve for scale normalization (c) and rotate the major axis to align with the x-axis for rotation normalization (d) simultaneously. Fourthly, we fix the starting point at the positive unit vector on the x-axis for starting point normalization (e). After these steps, we adjust the sign values for the x-symmetric and y-symmetric normalization (f) according to the 2nd order elliptic Fourier coefficients at last.

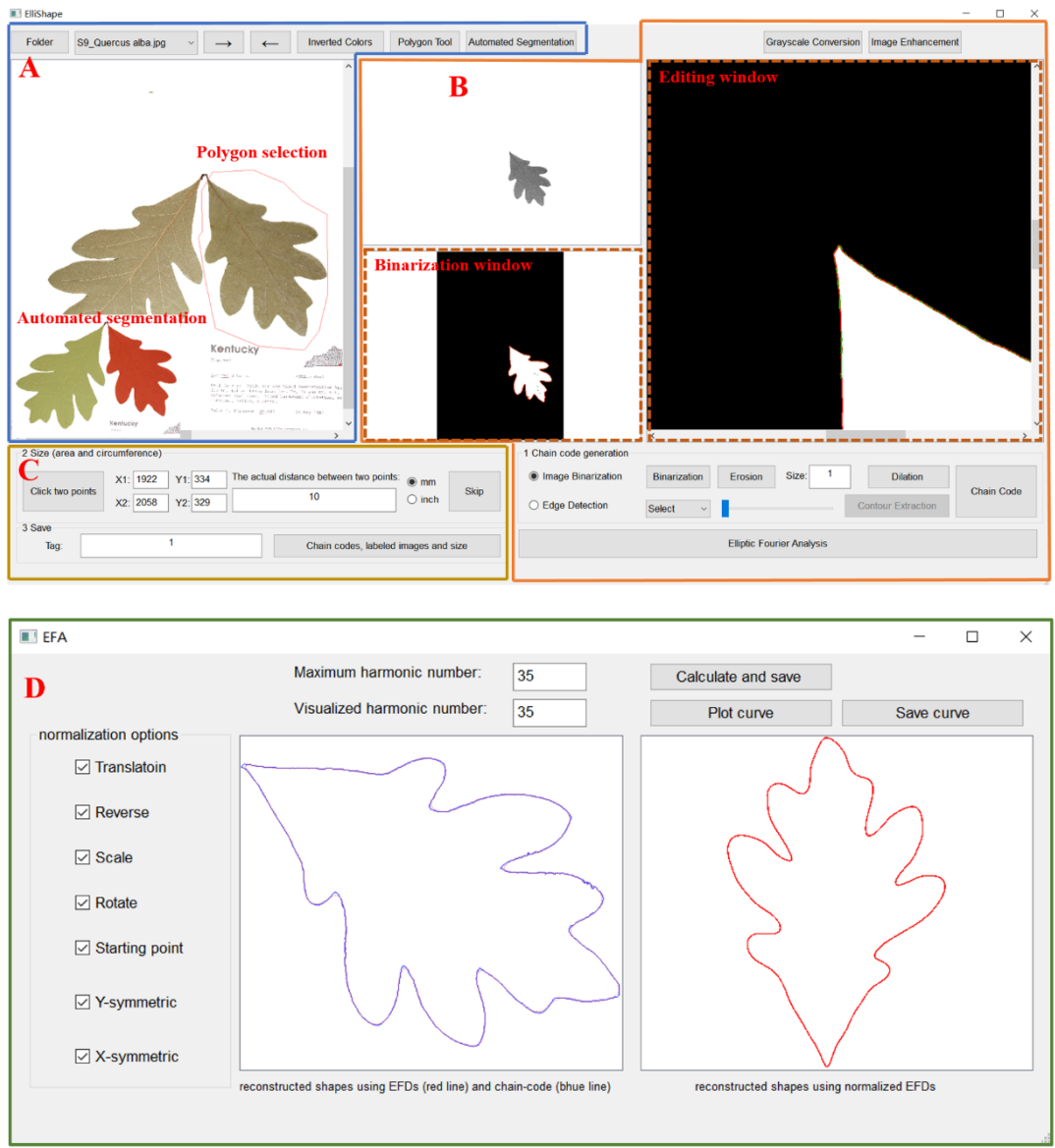

Figure 2. ElliShape software architecture, which consists of two main components: contour/outline extraction (A-C) and elliptic Fourier analysis (D). The contour/outline extraction component involves three primary functions: object selection (A), segmentation and chain code generation (B), as well as size calculation and saving (C). The second component (D) focuses on normalizing EFDs and visualizing reconstructed shapes.

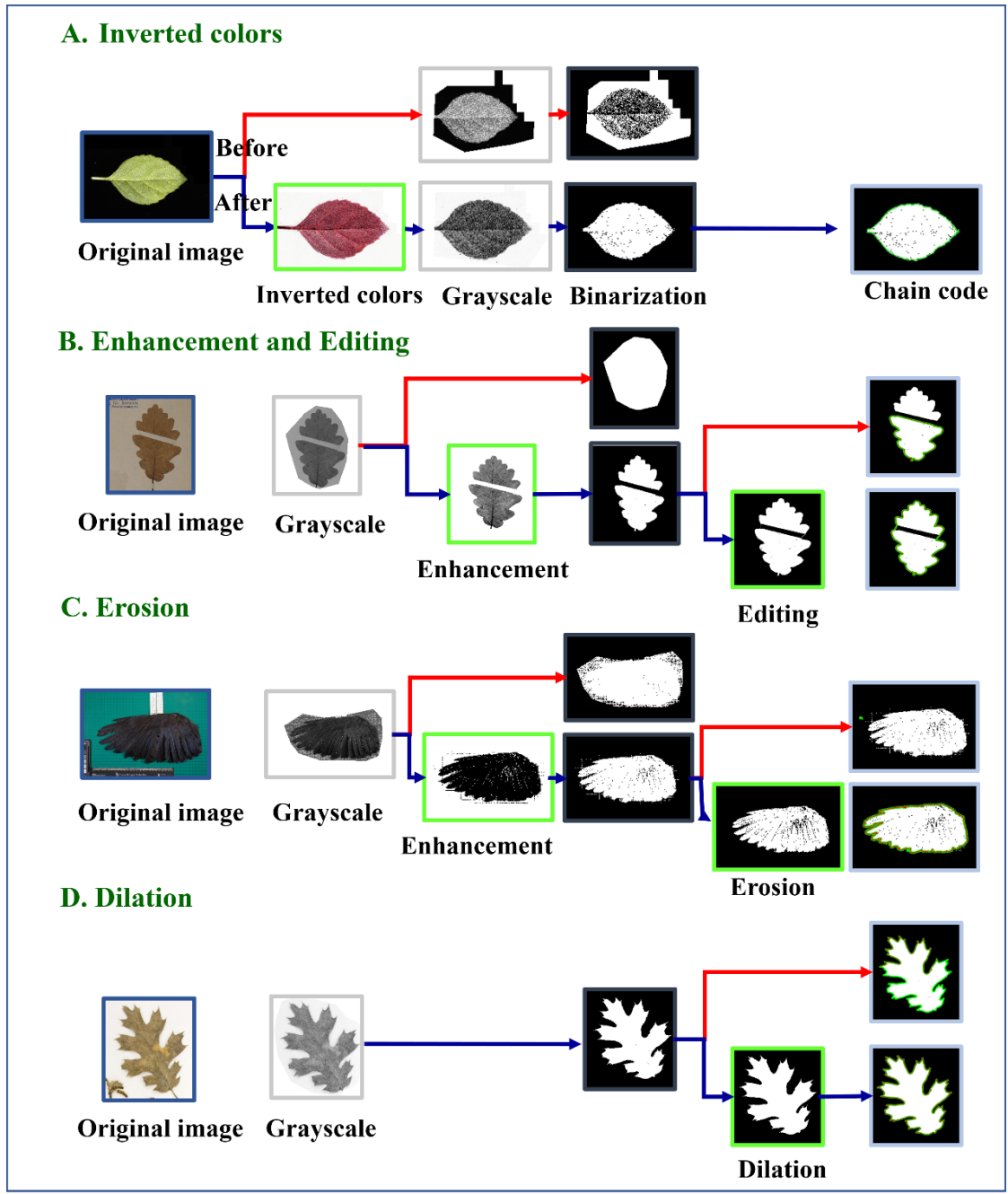

Figure 3. Contour/outline extraction results using various functions: (A) Inverted colors; (B) Image enhancement and contour editing; (C) Erosion operation; (D) Dilation operation.

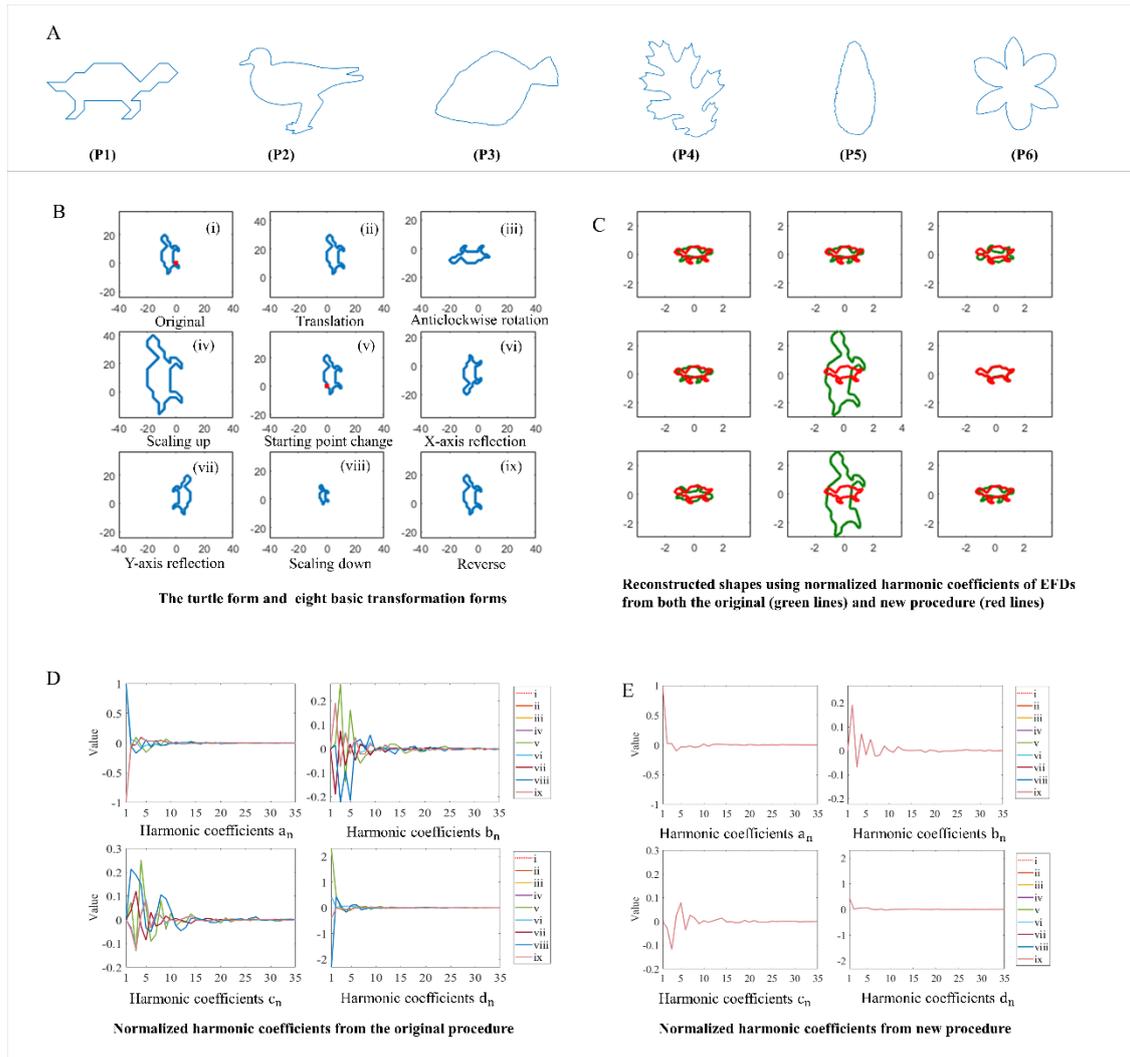

Figure 4. Comparison of elliptic Fourier analysis methods between the original and new procedures. (A), six-representative graphics, (B-E), results for typical turtle.

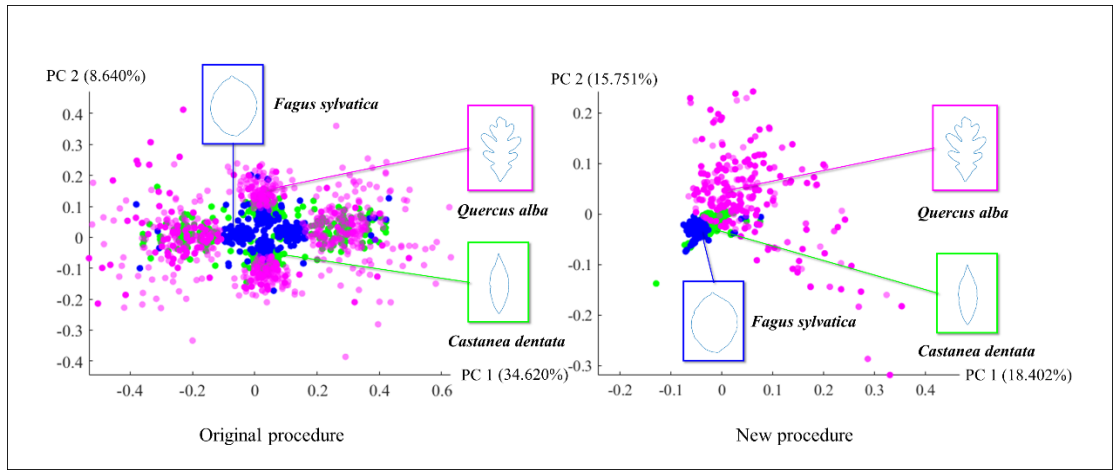

Figure 5. Comparison of principal component analysis (PCA) on the normalized EFDs from 319 leaves of three Fagaceae species using the original and new procedures. Each leaf contour/outline and its eight basic transformations serve as input data for the PCA analysis.

**Supporting information**

**Appendix S1 The approach for obtaining the elliptic Fourier descriptors. (A) Basic line segments of digital curve, (B) Example of a curve represented by chain code, (C) Elliptic Fourier Descriptors coefficients, (D) Reconstructed shape of the original curve using different order of EFDs.**

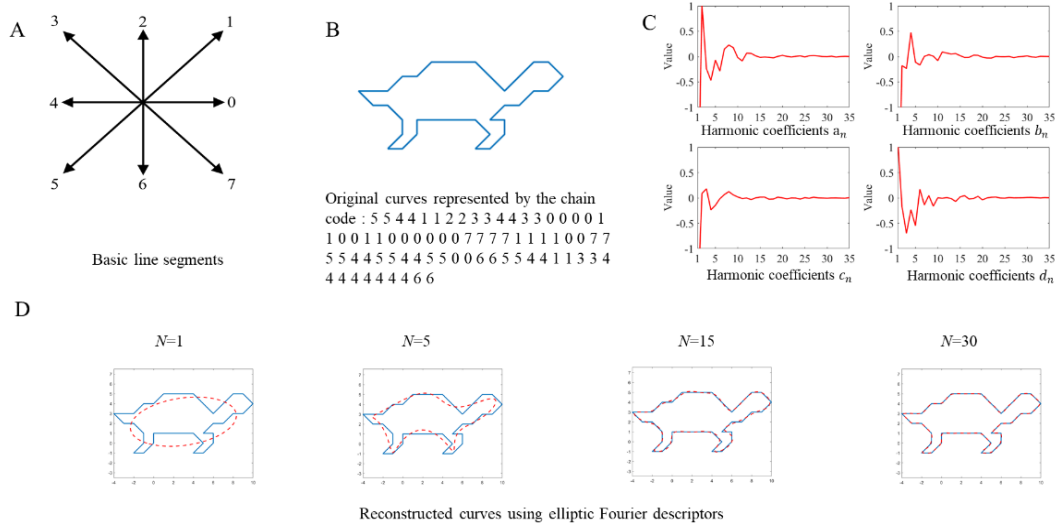

Reconstructed curves using elliptic Fourier descriptors

**Appendix S2 ElliShape User Guide**

1. **Introduction**

The elliptic Fourier analysis (EFA), originally proposed by Kuhl and Giardina in 1982, extracts arbitrary two-dimensional curves into elliptic Fourier descriptors (EFDs) for object recognition in images. EFA is a powerful mathematical tool widely utilized in various fields such as plant science, marine biology, evolutionary biology and anthropology for shape analysis, classification, reconstruction, and modeling.

Geometric morphometric procedures employing EFA include contour extraction, EFDs calculation, and EFD normalization. Despite it widespread use, challenges persist, including inefficiencies in outline extraction and limited dataset availability. EFD normalization, in particular, remains problematic, with existing methods unable to consistently produce homologous results across various basic transformations (Haines and Crampton 2000, Bonhomme et al., 2014, Wishkerman et al., 2018). Improving the computational efficiency of EFA for true EFD normalization and developing user-friendly software for organ annotation and geometric feature extraction from digital specimens are critical endeavors.

In this study, we introduced ElliShape, a user-friendly software designed to offer improved contour extraction, efficient EFDs calculation, and true EFD normalization across all basic contour transformations.

2. **Download and compile**

The software, developed in Python, requires Python 3.8 or later for compilation and execution. You can download it from the website: https://www.plantplus.cn/ElliShape. After downloading, extract all files from 'ElliShape.zip' into a folder named 'ElliShape'. To run the software (Fig. 1), double-click 'ElliShape.exe', no installation is required.

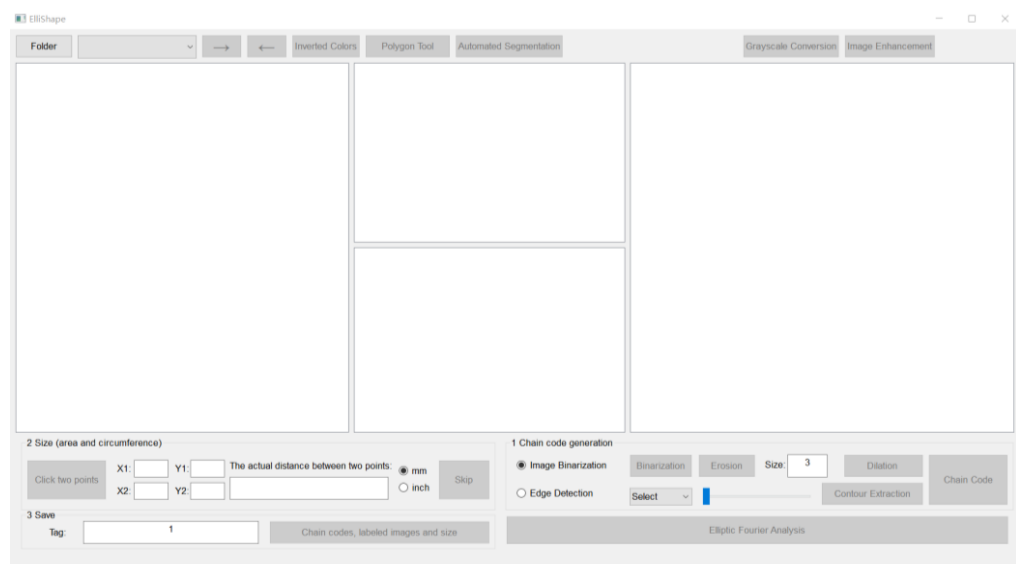

Figure 1. Software interface.

## 3. Functions

The software comprises two components: contour extraction and elliptic Fourier analysis.

### 3.1 Contour extraction page

Contour extraction involves four steps: manual target selection, automatic segmentation, manual correction of automatic contour outlining, and automatic chain code generation. Various functions address challenges of contour extraction during specimen preservation and digitization, such as incomplete edges due to white strips, low image contrast, and noises from fragile, overlapping, and damaged organs.

**Step 1: Image loading**

Click the 'Folders' button and choose the folder containing your image files. All file names with extensions '.jpg', '.png', '.tif', and '.bmp' will be listed. Choose the desired image file to load it into the program (Fig. 2). Use the 'Next' and 'Previous' buttons to navigate through the loaded images.

The test images can be found in the 'ElliShape' folder you downloaded.

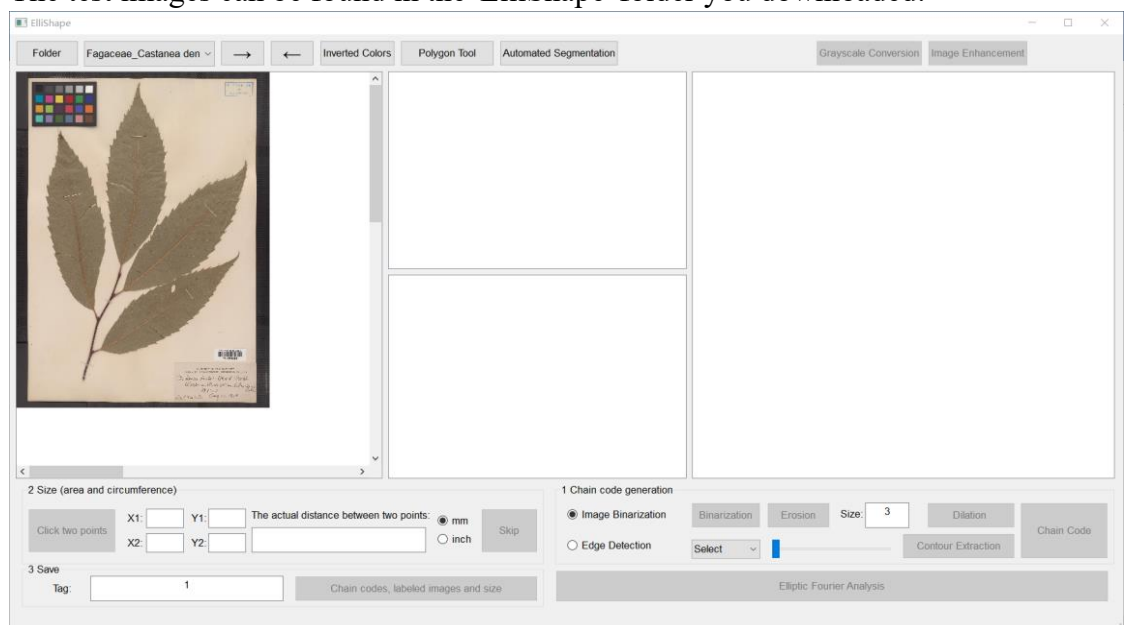

Figure 2. Image loading.

**Step 2: Color inverting**

If the foreground objects exhibit high grayscale values and distinct contrast with the background, the 'Inverted Colors' function is necessary. However, if they do not, it should be utilized.

An illustration of an image featuring a black background is shown in Fig. 3.

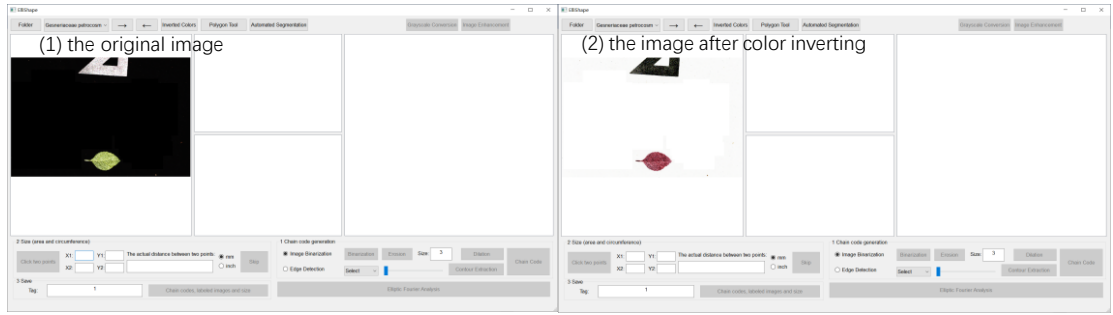

Figure 3. Color inverting.

**Step 3: Object selection**

Users can zoom in and out using the mouse scroll wheel and navigate left and right by dragging the slider to quickly locate the object (Fig. 4). Object selection is facilitated through two methods: automated segmentation and manual polygon selection.

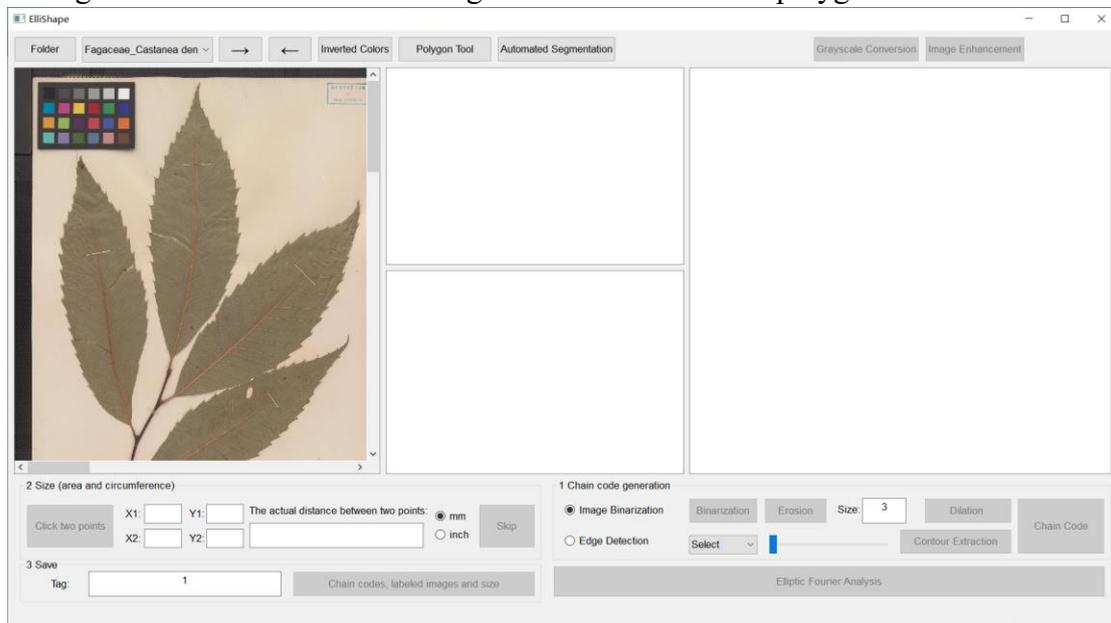

Figure 4. Object selection.

### Method 1: Object selection via "automated segmentation"

After selecting the object's center point in the left window, click the 'automated segmentation' button. The segmented object will appear in the lower middle and right windows as a binary image (Fig.5), enabling direct chain code generation without the need for "Step 4".

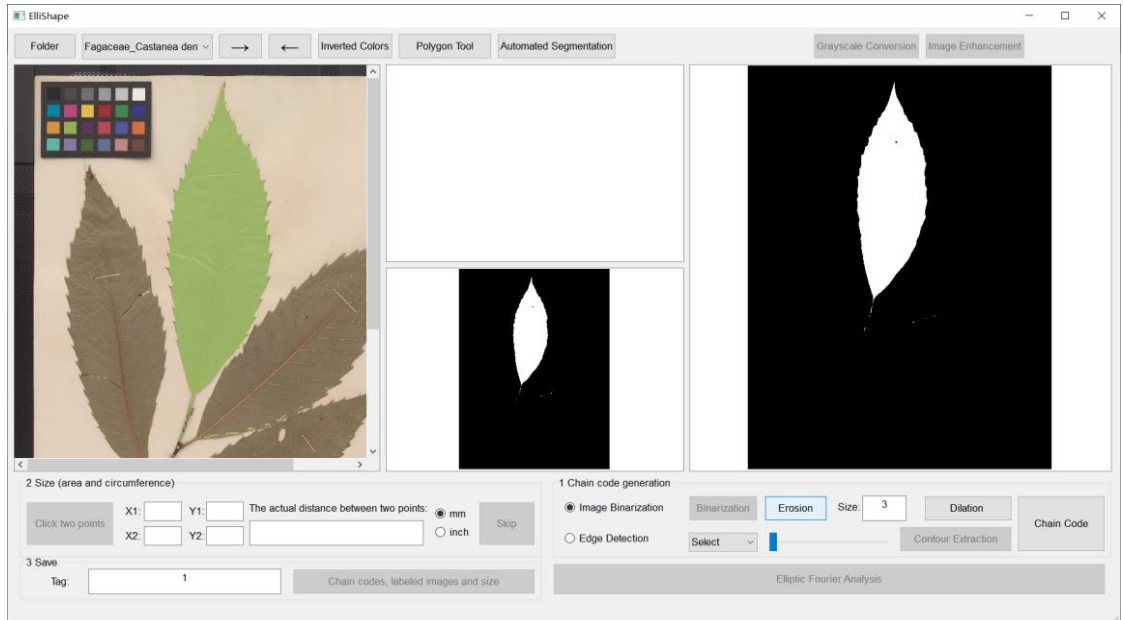

Figure 5. Object selection via "automated segmentation".

**Method 2: Object selection via "Polygon Tool"**

To select a region of interest (ROI), use the 'Polygon Tool' button. Left-click to create a a polygon around the leaf, connecting points. Right-clicking to close the polygon and display the selected object in the upper middle picture window (Fig. 6). To begin a new selection, right-clicking to end the current one.

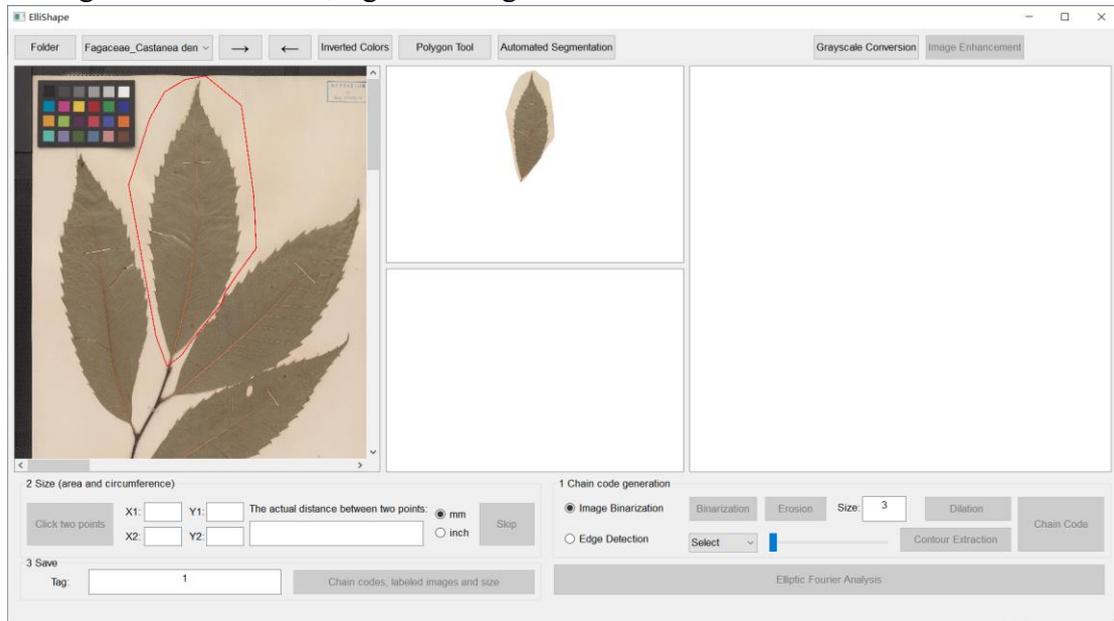

Figure 6. Object selection via "Polygon Tool".

**Step 4: Image enhancement and grayscale conversion**

If the foreground objects lack noticeable contrast with the background, use this step, as shown in Fig. 7.

Note: Clicking the 'Image Enhancement' button multiple times will superimpose the enhanced effect. To revert to the original grayscale image, click the 'Grayscale Conversion' button.

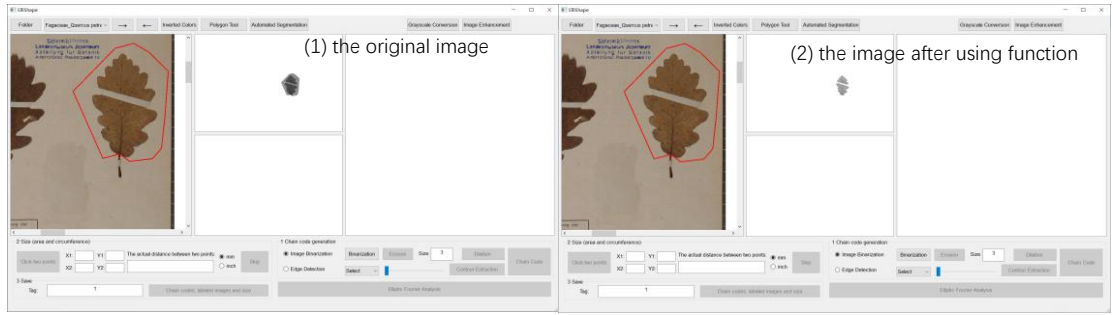

Figure 7. Image enhancement.

**Step 5: Chain code generation**

Two methods are available for obtaining the object contour: 'Image Binarization' or 'Edge Detection', selectable via radio buttons (Fig. 8). 'Image Binarization' is the default method for contour extraction.

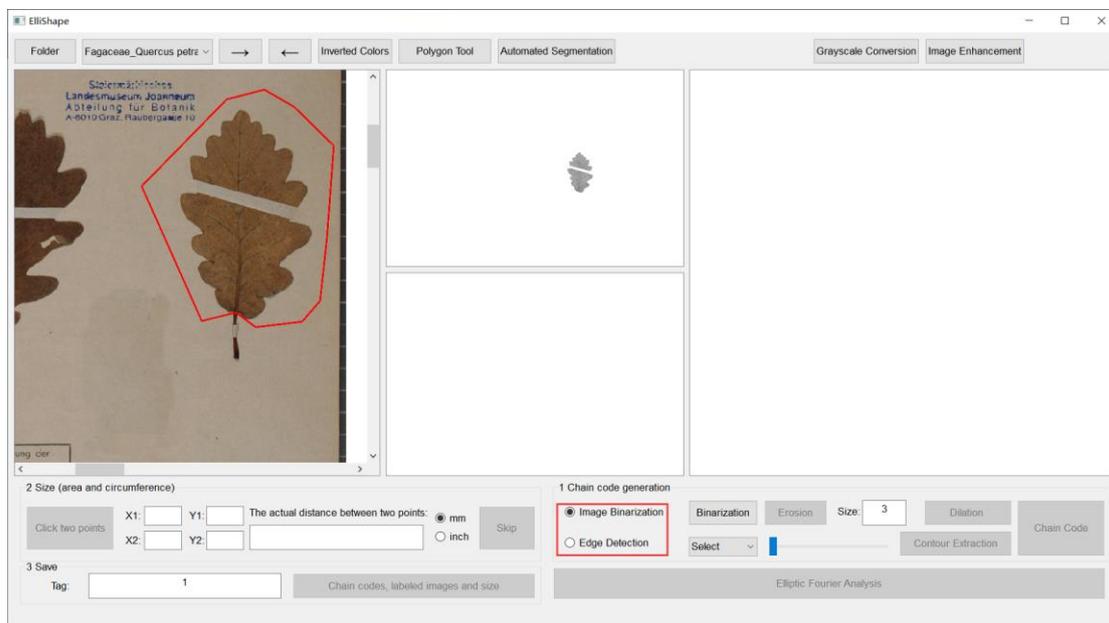

Figure 8. Two methods for obtaining the object contour.

**Method 1: Contour extraction via "Image binarization"**

Click the 'Binarization' button to convert the processed grayscale image into a binary image. The result will be displayed in the window below the middle section and in the enlarged view window on the right (Fig. 9).

To reduce noise in images with short chain code lengths (e.g. "the length is 2") , click the ' Erosion ' button (Fig. 10). Adjust the value in the box next to this button to control the size of the corrosion operation for desired results.

If the prompt indicates 'chain code is not close' and there are small gaps or holes in the target region hindering closed curve formation, click the 'Dilation' button to close the boundary (Fig. 11).

If the prompt indicates 'chain code is not close' and the editing window shows that a single pixel point causing interruption in the connecting line at the edge top, clicking the 'Corrosion' button or 'Dilation' button will close the boundary (Fig. 12). If your target object is broken, use the 'Editing' function in the right window. Zoom in using the mouse scroll wheel and pan the image, and draw white line to connect the broken parts by long-pressing the left mouse and dragging (Fig. 13). Conversely, dragging the right mouse button will draw black lines, breaking the connection.

Notes:

(1) The more dilations performed, the greater the distortion of the object, so use an appropriate number.

(2) Zoom in until one pixel is clearly visible before editing the image to avoid positional deviations (Fig. 13).

(3) Decide whether to use corrosion, dilation, and editing functions based on the specific circumstances.

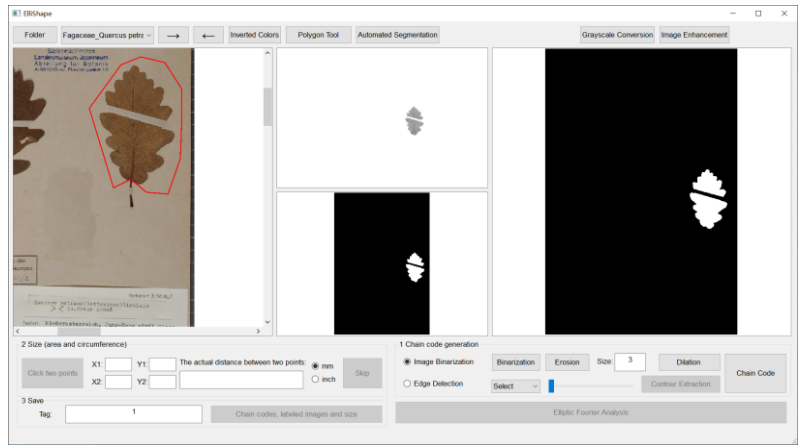

Figure 9. Binarization results.

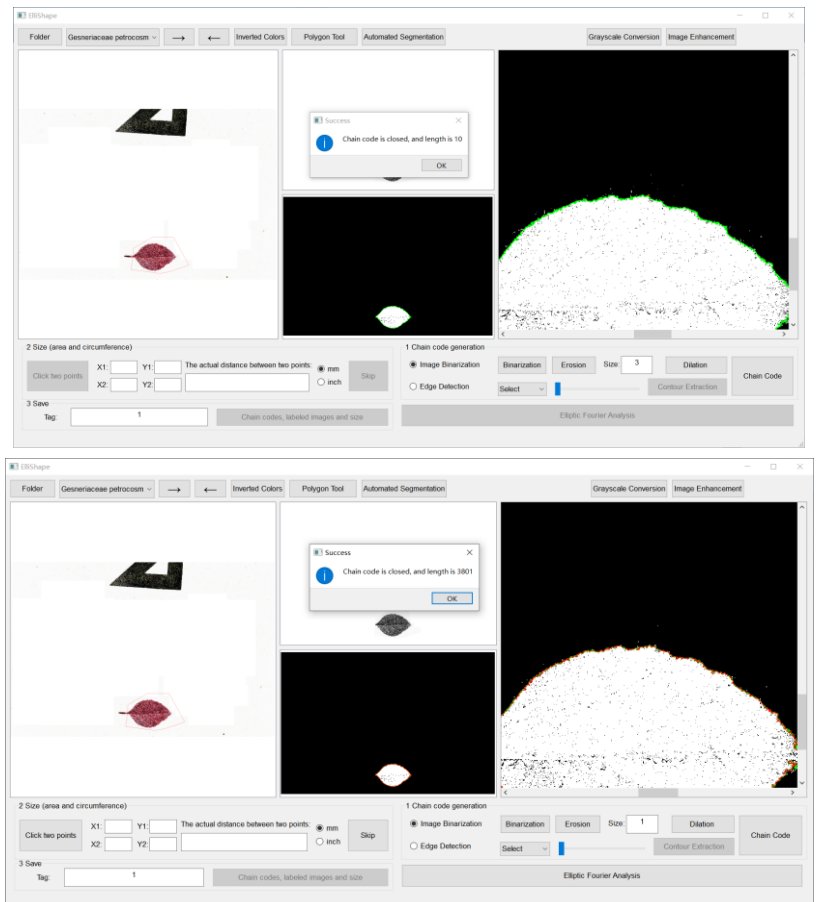

Figure 10 Erosion operation. When the prompt words indicate the short length of chain code (e.g. "the length is 2"), click the ' Erosion ' button and obtain the corrected boundary.

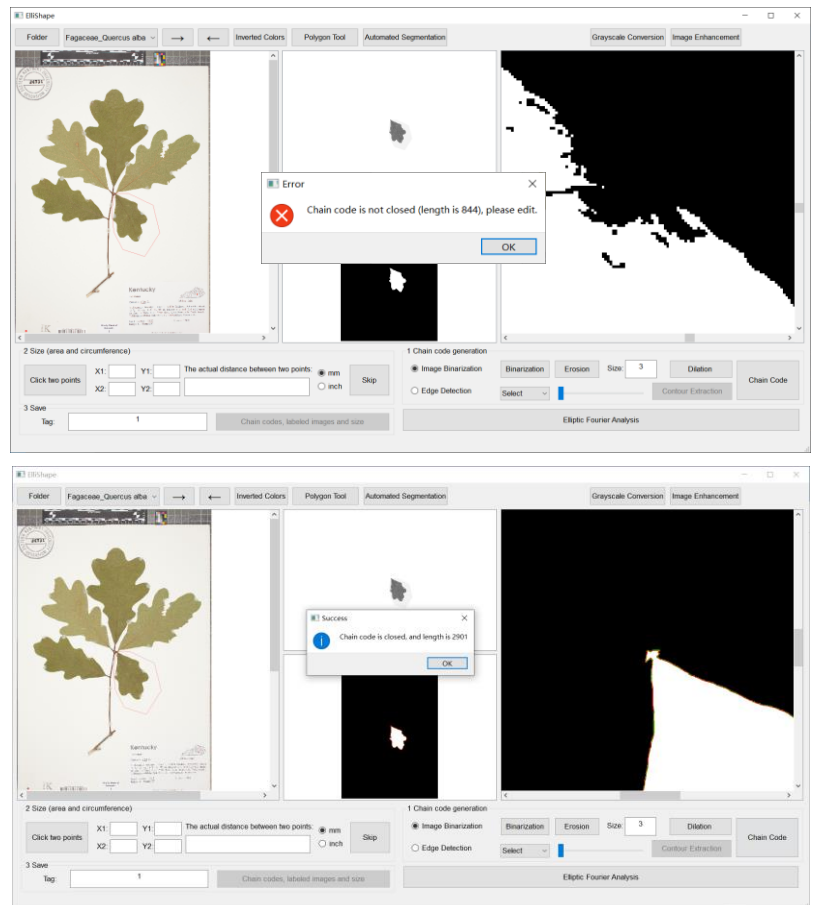

Figure 11. Dilation operation. When the prompt indicates 'chain code is not close' and there are small gaps or holes in the target region hindering closed curve formation, click the 'Dilation' button to close the boundary.

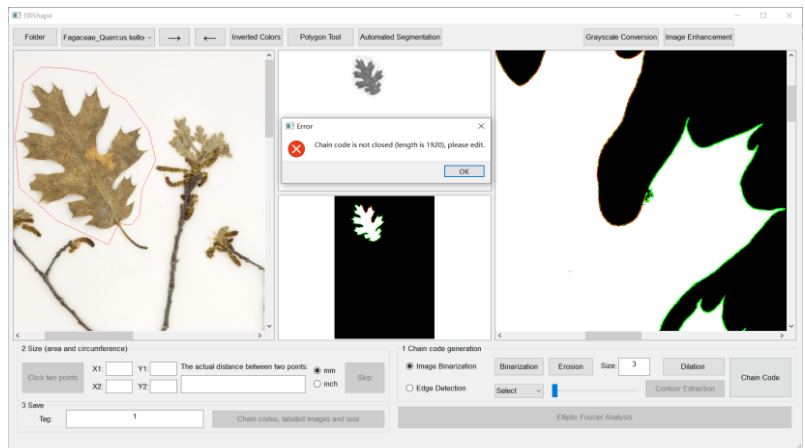
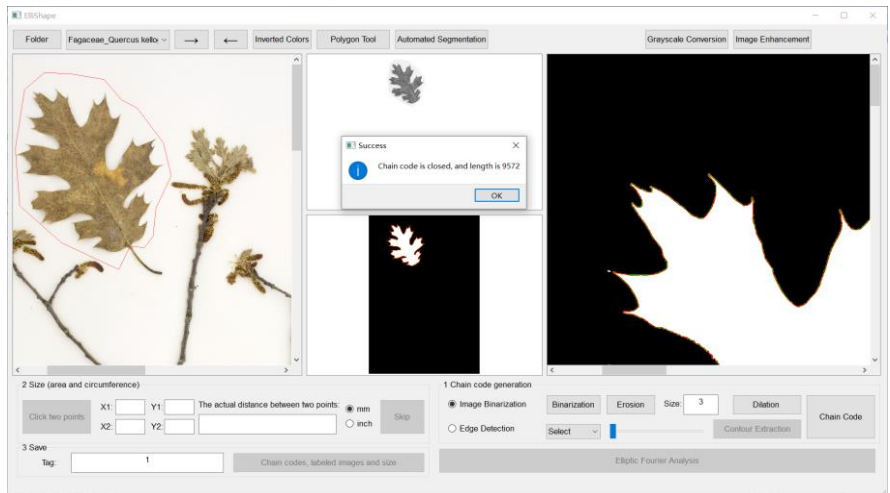

Figure 12. Dilation operation. When prompted that the 'chain code is not closed' and the editing window shows a single pixel point interrupting the connecting line at the top edge, clicking 'Dilation' button will close the gap in the boundary.

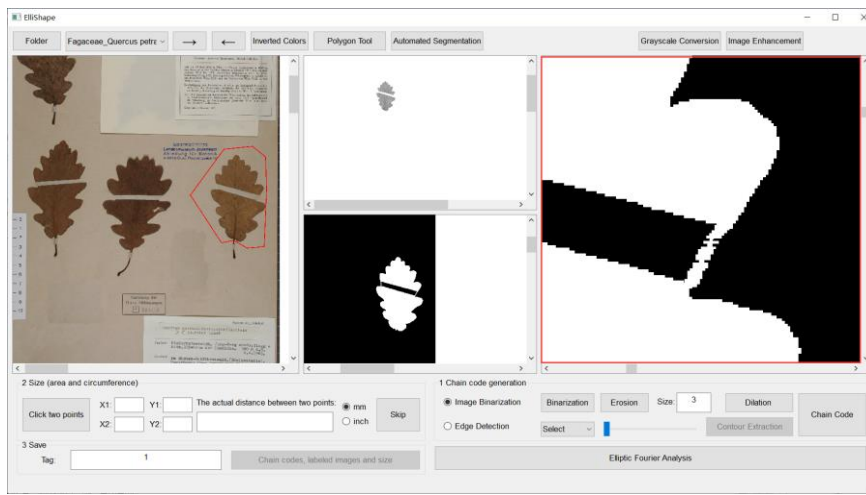

Figure 13. Editing operation. When target object is broken, draw white line to connect the broken parts by long-pressing the left mouse and dragging.

Click the 'Chain Code' button to initiate chain code extraction, with a message box confirming its success (Fig. 14). The correctness and closure of the chain code are determined by its length and the presence of a red line in the middle right window. If the chain code is not closed, the break location is enlarged in the editing window for user edits, followed by clicking the 'Chain Code' button again until the result is correct.

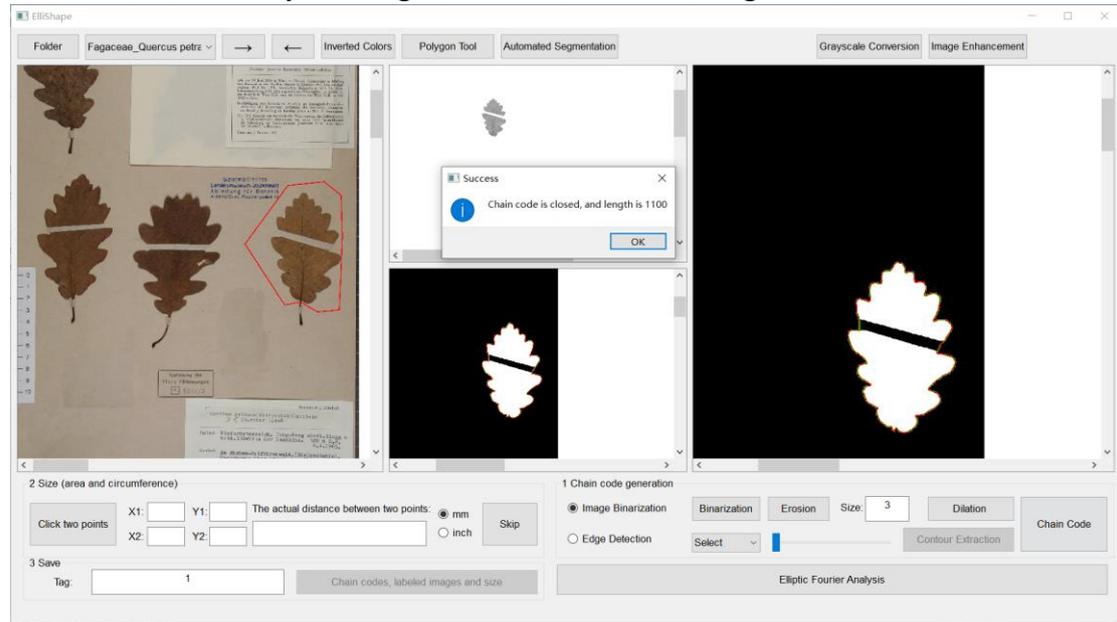

Figure 14. Successful chain code extraction.

**Method 2: Contour extraction via "Edge detection"**

Various edge detection methods, including Canny, Sobel, Prewitt, and Roberts operators, as well as the log and zero-cross detectors, are available in the popup menu (Fig. 15). Select the desired method adjust the threshold using the slider. A lower threshold reveals more details, while a higher threshold detects fewer details. Optimal thresholding is achieved when clear edge contours are visible (Fig. 16).

In the editing window, users can connect disjointed contours and correct any contour errors (Fig. 17). Click the 'Chain Code' button to obtain the correct edge contour (green line) and chain code (red line) as shown in Fig. 18.

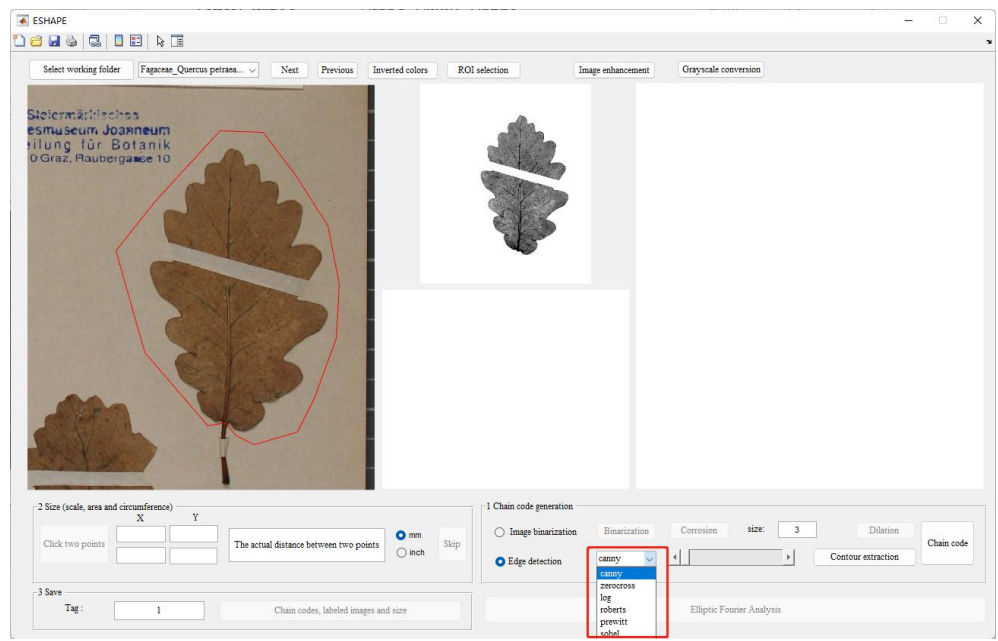

Figure 15. Edge detection method selection.

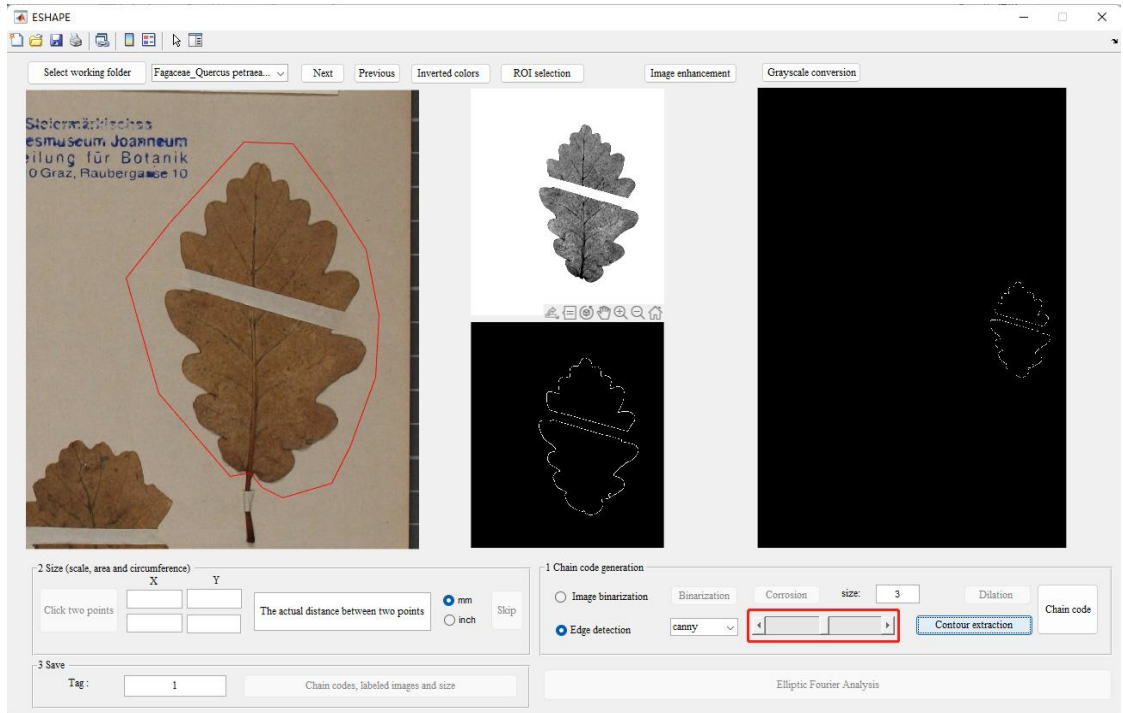

Figure 16. Threshold adjustment.

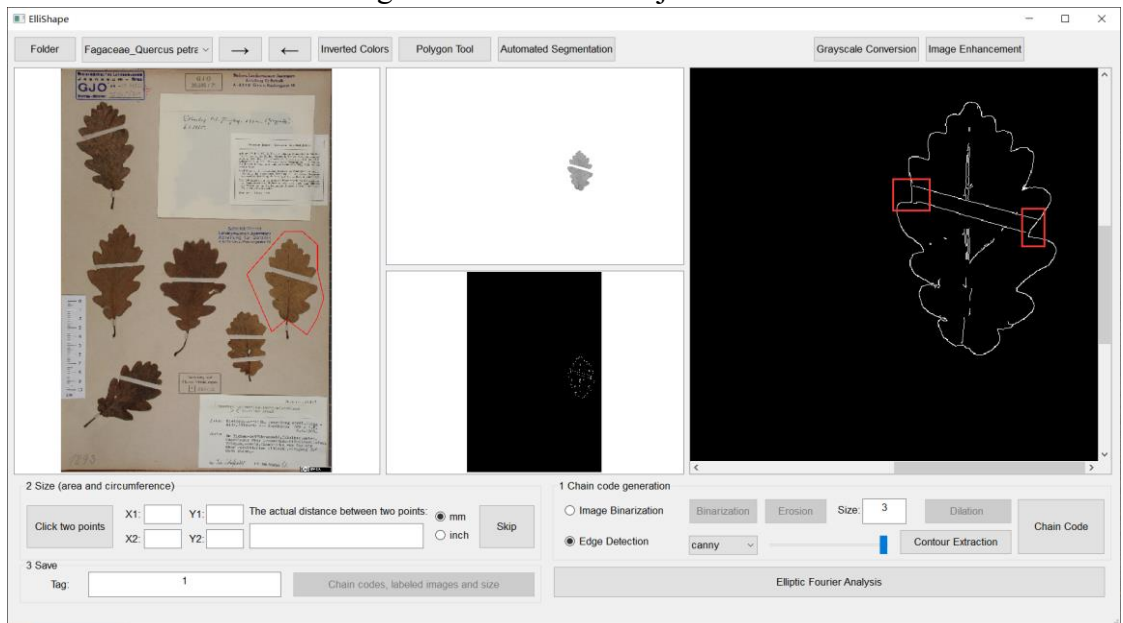

Figure 17. Contour connection.

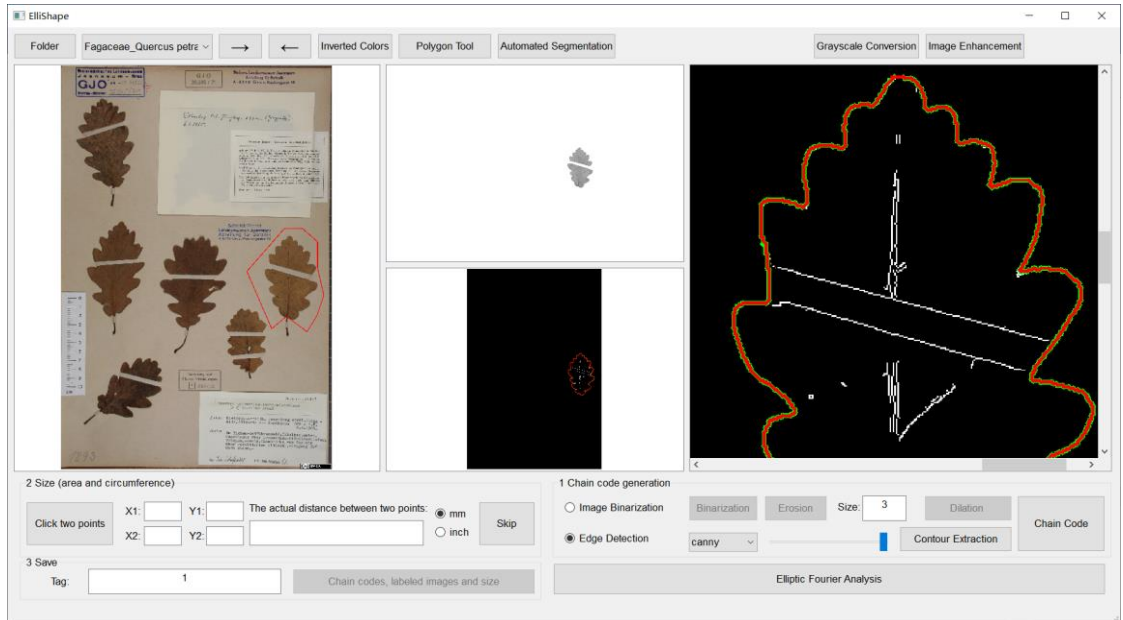

Figure 18. Successful chain code extraction.

**Step 6: Size calculation and saving**

If a ruler is present, zoom in on the image in the left window to the ruler's location. Click the 'Click Two Points' button and select a starting point and an endpoint by left-clicking. The coordinates of the two points will be displayed in the 'X' and 'Y' text boxes. Enter the actual distance between the points in millimeters in the right text box, for example, input '10' (Fig. 19). Choose the measurement units as 'mm' or 'inch'. If a ruler is absent, click the 'Skip' button. Provide a label for the object, then click the 'Chain code, Labeled images, and size' button to save the outputs with names appended with the user-defined tag. These outputs include:

Boundary coordinates: input file name_user-defined tag_b.txt
Chain code: input file name_user-defined tag_c.txt
Size: input file name_user-defined tag_info.xlsx, Sheet 1
Labeled images: input file name_user-defined tag.png

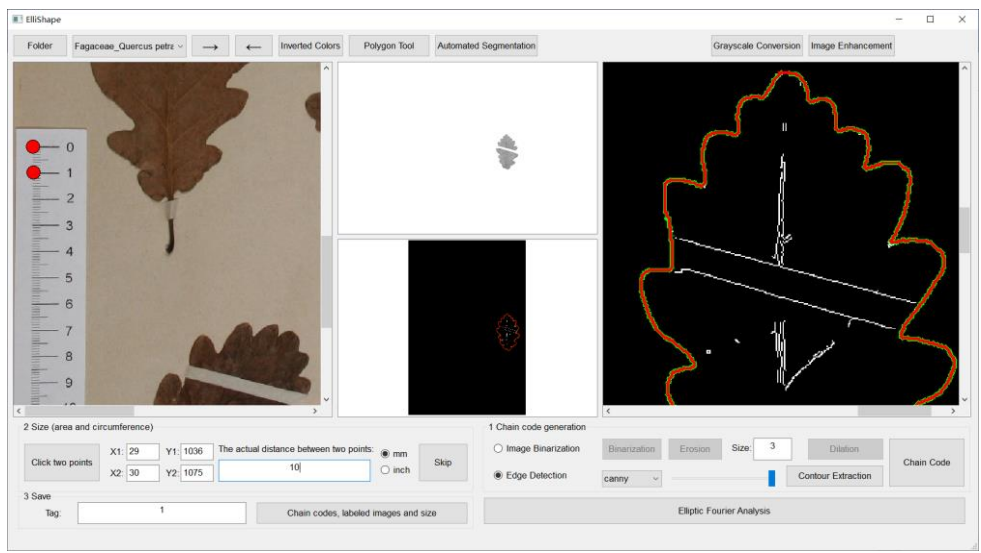

Figure 19. Size calculation.

## 3.2 Ellipse Fourier Analysis page

Click the 'Elliptic Fourier Analysis' button to access a new page for obtaining normalized EFD data and visualizing reconstructed shapes (Fig. 20).

Use the 'Calculate and Save' button to save the EFD data. Users can input any integer as the number of harmonic coefficients, with the default value set at 35 (Fig.21).

Adjust the visualization number of the EFD, ensuring it is less than or equal to the maximum, by clicking the 'Plot Curve' button. The result is displayed in Fig. 22. Click the 'Save Curve' button to choose the path for saving the curve (Fig. 23).

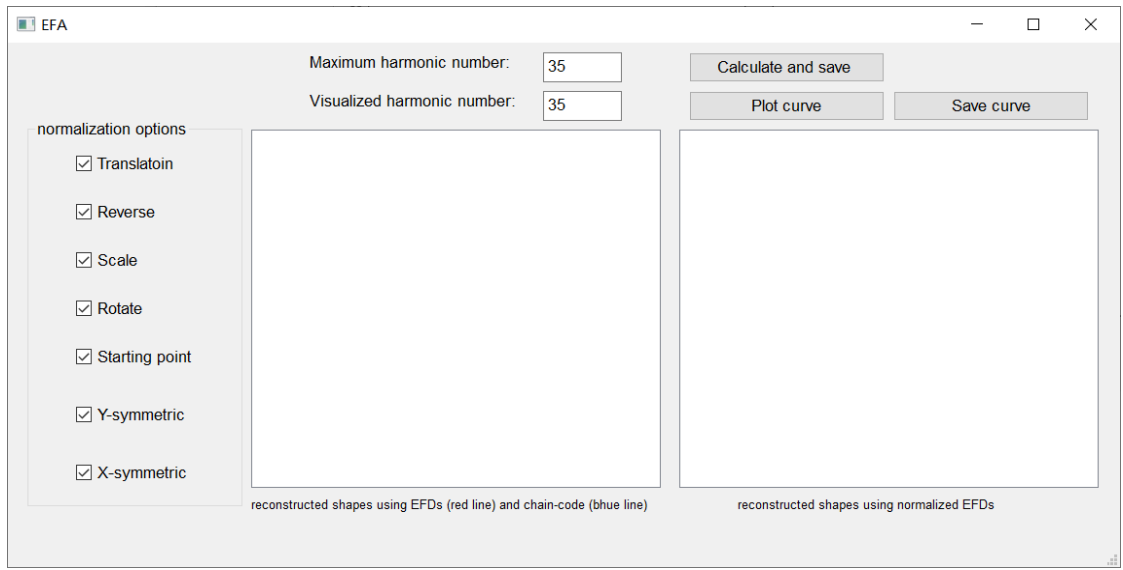

Figure 20. EFA interface.

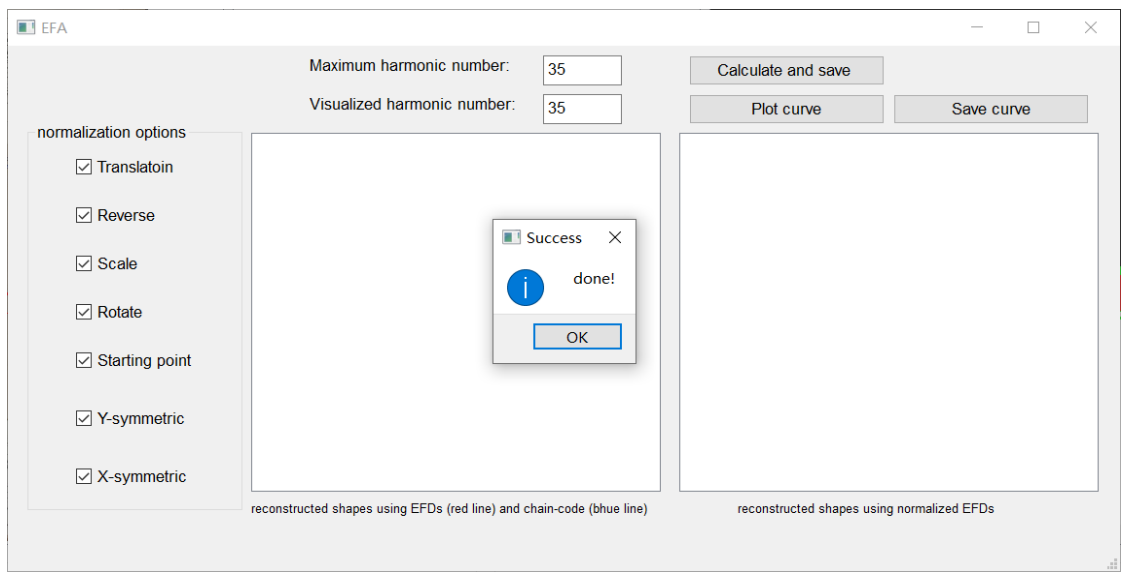

Figure 21. EFD calculation.

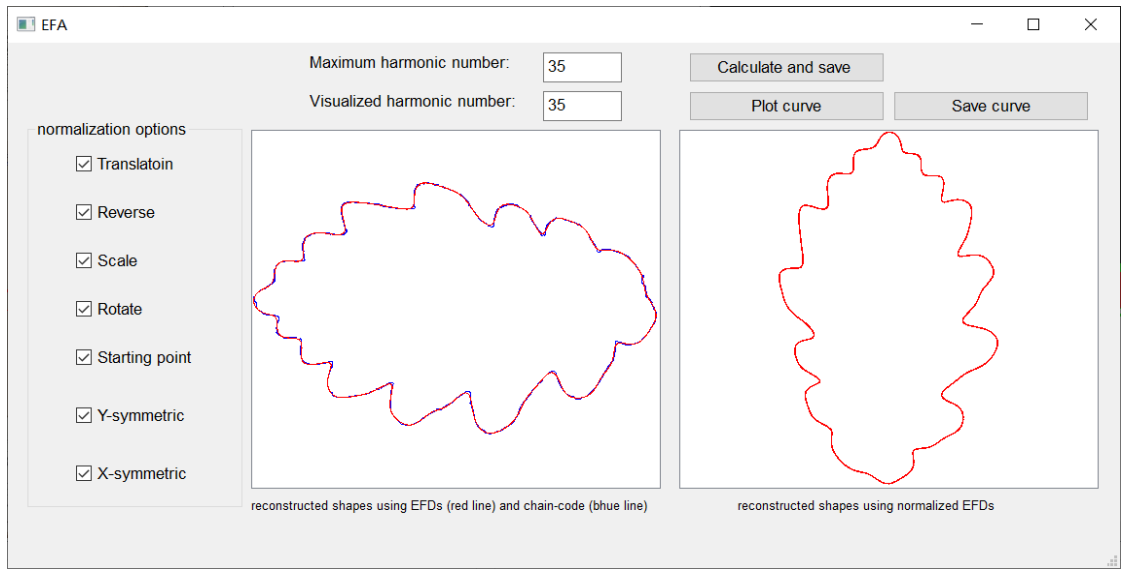

Figure 22. Curve reconstruction.

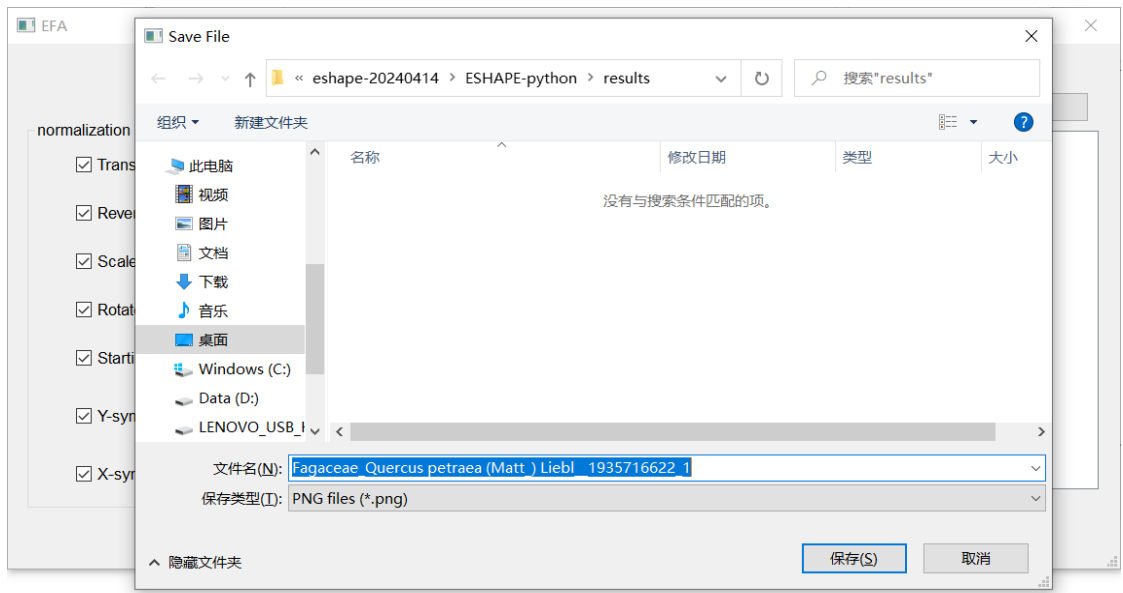

Figure 23. Curve saving.

**Appendix S3 The overall comparisons of EFD normalization between existing methods and our procedure**

| Software | original sample(i) | translation(ii) | anticlockwise rotation(iii) | scaling up(iv) | started with different starting point(v) | x-symmetric(vi) | y-symmetric(vii) | scaling down(viii) | reversed(ix) |
|---|---|---|---|---|---|---|---|---|---|
| original | P1 | √ | × | √ | √ | × | × | × | × |
| | P2 | √ | √ | √ | × | × | × | √ | × |
| | P3 | √ | × | √ | √ | × | × | √ | × |
| | P4 | √ | × | √ | × | × | × | √ | × |
| | P5 | √ | × | √ | × | √ | × | √ | × |
| | P6 | √ | × | √ | × | × | × | √ | × |
| SHAPE 1 (based on the First Harmonic) | P1 | √ | √ | √ | √ | × | × | √ | × |
| | P2 | √ | √ | √ | √ | × | × | √ | × |
| | P3 | √ | √ | √ | √ | × | × | √ | × |
| | P4 | √ | √ | √ | × | × | × | √ | × |
| | P5 | √ | √ | √ | × | × | × | √ | × |
| | P6 | √ | √ | √ | × | × | × | √ | × |
| SHAPE 2 (based on the Longest radius) | P1 | √ | √ | √ | √ | × | × | × | × |
| | P2 | √ | √ | √ | √ | × | × | √ | × |
| | P3 | √ | √ | √ | √ | × | × | √ | × |
| | P4 | √ | √ | √ | √ | × | × | √ | × |
| | P5 | √ | √ | √ | √ | × | × | √ | × |
| | P6 | √ | √ | √ | × | × | × | √ | × |
| Momocs | P1 | √ | √ | √ | √ | × | × | √ | × |
| | P2 | √ | √ | √ | × | × | × | √ | × |
| | P3 | √ | √ | √ | √ | × | × | √ | × |
| | P4 | √ | √ | √ | × | × | × | √ | × |
| | P5 | √ | √ | √ | × | × | × | √ | × |
| | P6 | √ | √ | √ | × | × | × | √ | × |
| MASS | P1 | √ | √ | √ | × | × | × | × | × |
| | P2 | √ | × | √ | × | × | × | √ | × |
| | P3 | √ | √ | √ | × | × | × | √ | × |
| | P4 | √ | √ | √ | × | × | × | √ | × |
| | P5 | √ | √ | √ | × | × | × | √ | × |
| | P6 | √ | × | √ | × | × | × | √ | × |
| our new procedure | P1 | √ | √ | √ | √ | √ | √ | √ | √ |
| | P2 | √ | √ | √ | √ | √ | √ | √ | √ |
| | P3 | √ | √ | √ | √ | √ | √ | √ | √ |
| | P4 | √ | √ | √ | √ | √ | √ | √ | √ |
| | P5 | √ | √ | √ | √ | √ | √ | √ | √ |
| | P6 | √ | √ | √ | √ | √ | √ | √ | √ |

**Appendix S4 Comparison of elliptic Fourier analysis methods between the original and new procedures for graphics P2-P6.**

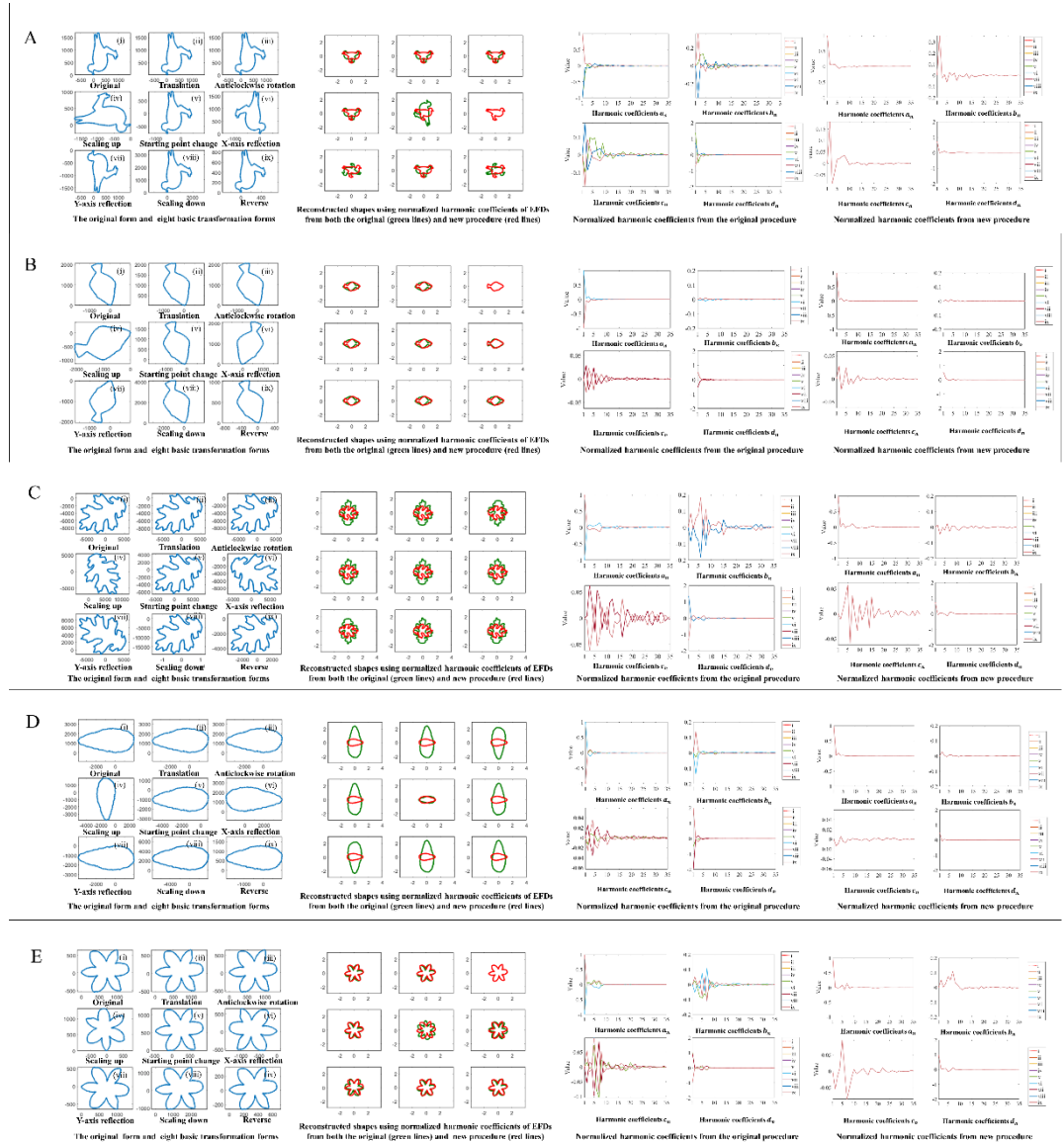

**Appendix S5** Comparison of principal component analysis (PCA) on the normalized EFDs from 207 cones three *Pinus* species using the original and new procedures. Each cone contour/outline and its eight basic transformations serve as input data.

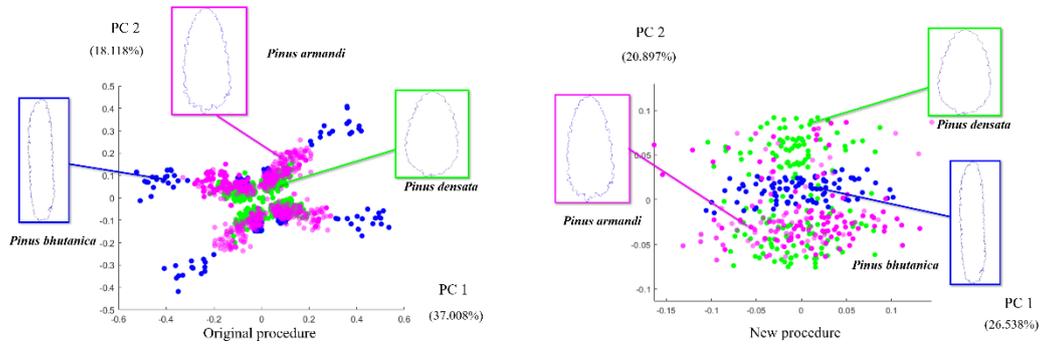

**Appendix S6** Overall comparison of contour/outline extraction between existing methods and our procedure.